\documentclass{article} % For LaTeX2e
\usepackage{iclr2024_conference,times}

\usepackage{amsmath,amsfonts,bm}

\def\eqref#1{equation~\ref{#1}}
\def\1{\bm{1}}

\DeclareMathAlphabet{\mathsfit}{\encodingdefault}{\sfdefault}{m}{sl}
\SetMathAlphabet{\mathsfit}{bold}{\encodingdefault}{\sfdefault}{bx}{n}

\usepackage{hyperref}
\usepackage{url}
\usepackage{graphicx}
\usepackage[utf8]{inputenc} % allow utf-8 input
\usepackage[T1]{fontenc}    % use 8-bit T1 fonts
\usepackage{hyperref}       % hyperlinks
\usepackage{url}            % simple URL typesetting
\usepackage{booktabs}       % professional-quality tables
\usepackage{amsfonts}       % blackboard math symbols
\usepackage{nicefrac}       % compact symbols for 1/2, etc.
\usepackage{microtype}      % microtypography
\usepackage{array}
\usepackage{multirow}
\definecolor{cvprblue}{rgb}{0.21,0.49,0.74}
\usepackage{xcolor}

\title{Data Alignment for Zero-Shot Concept Generation in Dermatology
AI}

\author{Soham Gadgil \thanks{Equal contribuation}      \   , Mahtab Bigverdi \footnotemark[1] \\
Paul G. Allen School of Computer Science and Engineering\\
University of Washington\\
\texttt{\{sgadgil,mahtab\}@cs.washington.edu} \\
}

\iclrfinalcopy % Uncomment for camera-ready version, but NOT for submission.
\begin{document}
\maketitle

\begin{abstract}
  AI in dermatology is evolving at a rapid pace but the major limitation to training trustworthy classifiers is the scarcity of data with ground-truth concept level labels, which are meta-labels semantically meaningful to humans \citep{li2019artificial}. Foundation models like CLIP \citep{radford2021learning} providing zero-shot capabilities can help alleviate this challenge by leveraging vast amounts of image-caption pairs available on the internet. CLIP can be fine-tuned using domain specific image-caption pairs to improve classification performance. However, CLIP's pre-training data is not well-aligned with the medical jargon that clinicians use to perform diagnoses. The development of large language models (LLMs) in recent years has led to the possibility of leveraging the expressive nature of these models to generate rich text. Our goal is to use these models to generate caption text that aligns well with both the clinical lexicon and with the natural human language used in CLIP's pre-training data. Starting with captions used for images in PubMed articles \citep{kim2023fostering}, we extend them by passing the raw captions through an LLM fine-tuned on the field's several textbooks. We find that using captions generated by an expressive fine-tuned LLM like GPT-3.5 improves downstream zero-shot concept classification performance. 
\end{abstract}

\section{Introduction}
In dermatology, for performing a diagnosis, dermatologists often use concepts, which refer to a clinical lexicon that is used to describe skin disease findings in the dermoscopic images. For example, Melanoma is often associated with the ABCDE rule including asymmetry, border, color, diameter and evolving \citep{duarte2021clinical}. Thus, learning these concepts from an image can aid in providing diagnostic explanations and building classifiers which are explainable. However, obtaining these concept labels for dermatology is a difficult and time-consuming task since only well-trained dermatologists can accurately describe skin diseases. There are datasets \citep{codella2018skin, groh2021evaluating} which have high-quality dermoscopic images, but they are either devoid of manual labels, not inclusive of all concepts, or have very limited samples for some concepts. 

There have been many advances in fully-supervised learning for medical image classification spanning multiple domains \citep{yadav2019deep, islam2020deep, li2014medical}. However, the same progress has not been achieved in dermatology image analysis due to limited availability of high-quality images with expert annotations. Recently introduced methods like CLIP provide avenues to perform zero-shot classification without the need of labeled datasets. Prior works like MONET \citep{kim2023fostering} leverage image-caption pairs from PubMed articles and medical textbooks to fine-tune CLIP models for dermatology. However, the captions used in these academic sources contain medical terms which are not aligned with the pre-training data of CLIP, which includes image-caption pairs found on the internet. We posit that LLMs like GPT variants can be effectively used to model natural human language. Our contributions include (i) using LLMs for data generation by extending the original captions to align them with CLIP's pre-training data and improve downstream performance on zero-shot concept classification, (ii) demonstrating that these LLMs can be further fine-tuned on the field's textbooks to improve their expressiveness.
% The pipeline for this work is illustrated in Figure \ref{pipeline}.\\

\section{Datasets}
\paragraph{Textbooks}
With the advent of LLMs, many open-source and closed-source LLM models pre-trained on vast amounts of open internet text data are available. Although, for a specific task like this work, an improved and more informative text in the dermatology field is required that some of these pre-trained models cannot provide. Therefore, fine-tuning a LLM on the desired text set is a crucial solution to this problem. Dermatology textbooks are a good option for fulfilling this requirement. We chose four books for this purpose:  Differential Diagnosis In Dermatology \citep{ashton2021differential}, General Dermatology \citep{english2007general}, Top 50 Dermatology Case Studies for Primary Care \citep{reich2017top}, and Handbook of Dermoscopy \citep{malvehy2006handbook}. We used the text from these textbooks to generate prompt and completion pairs for fine-tuning the LLM models as described in section \ref{preprocess}.

\paragraph{Evaluation Dataset} To evaluate the trained CLIP model for zero-shot concept classification, we used the SKINCON dataset \citep{daneshjou2022skincon}. SKINCON includes 3230 images from the Fitzpatrick 17k skin disease dataset \citep{groh2021evaluating}, densely annotated with 48 clinical concepts, 22 of which have at least 50 images representing the concept. The concepts used were chosen by two dermatologists considering the clinical descriptor terms used to describe skin lesions, such as "plaque", "scale", and "erosion" to name a few. The list of concepts was based on the clinical lexicon
used by dermatologists to describe skin lesions and was developed with consultation of the terms listed in one of the most widely used dermatology textbooks - Dermatology \citep{bolognia2012dermatology}.

\section{Methods}
\subsection{Exploratory Analysis}
For training CLIP, the captions need to be tokenized using the CLIP tokenizer before the contrastive learning procedure. All CLIP models use 77 as the maximum tokenized context length, either padding or truncating the caption if it is below or above that length respectively.

Since we were restricted to 77 as the maximum number of tokens, we first did an exploratory analysis of the tokenized lengths of the original 44314 captions obtained from the PubMed articles utilizing the scripts provided in \cite{kim2023fostering}. This would give us an intuition of how many tokens were available for extending the caption for alignment. Table \ref{cap_len_stats} (Appendix \ref{sec:tables}) shows the statistics of the tokenized captions. The mean length of captions is $\sim$ 35 which shows that most of the captions are short and do not exceed the maximum token length of 77. 75\% of the captions have a token length of less than 51 which indicates that a majority of captions do have additional tokens available to be extended and improved. There are $\sim$ 13\% captions which have been truncated at the max token length of 77, still leaving around $\sim$ 38000 captions that can be improved using LLMs. 

% \begin{table}[h]
%   \caption{Tokenized captions length statistics}
%   \label{cap_len_stats}
%   \centering
%   \begin{tabular}{||  m{6cm} |  m{1cm}  ||}
%  \hline
%  Statistic & Value  \\ [0.5ex] 
%  \hline\hline
%  Mean & 35.394 \\ 
%  \hline
% Standard Deviation & 22.801\\
%  \hline
 
%  Median & 28 \\
%  \hline
% Minimum & 3 \\
%  \hline
%  Maximum & 77 \\
%  \hline
%  \hline
% \end{tabular}

% \end{table}

% \begin{figure}[h]
%     \centering
%   \includegraphics[width=0.49\textwidth]{images/tokenized_caption_lengths_hist.png}
%   \caption{ Distribution of tokenized caption lengths}
%   \label{cap_len}
% \end{figure}

\subsection{Data preprocessing} \label{preprocess}
Fine-tuning data for an LLM needs to be in the form of prompt-completion pairs. It meant for a specific prompt, we needed to define the ideal completion that we expect the model to output. Naively, these would be the sentences that follow a given prompt in the text. However, the whole of the raw text from the books could have misleading phrases and sentences, so applying some preprocessing strategies was essential for fine-tuning data preparation.

\begin{figure}[h]
 
  \includegraphics[width=15cm]{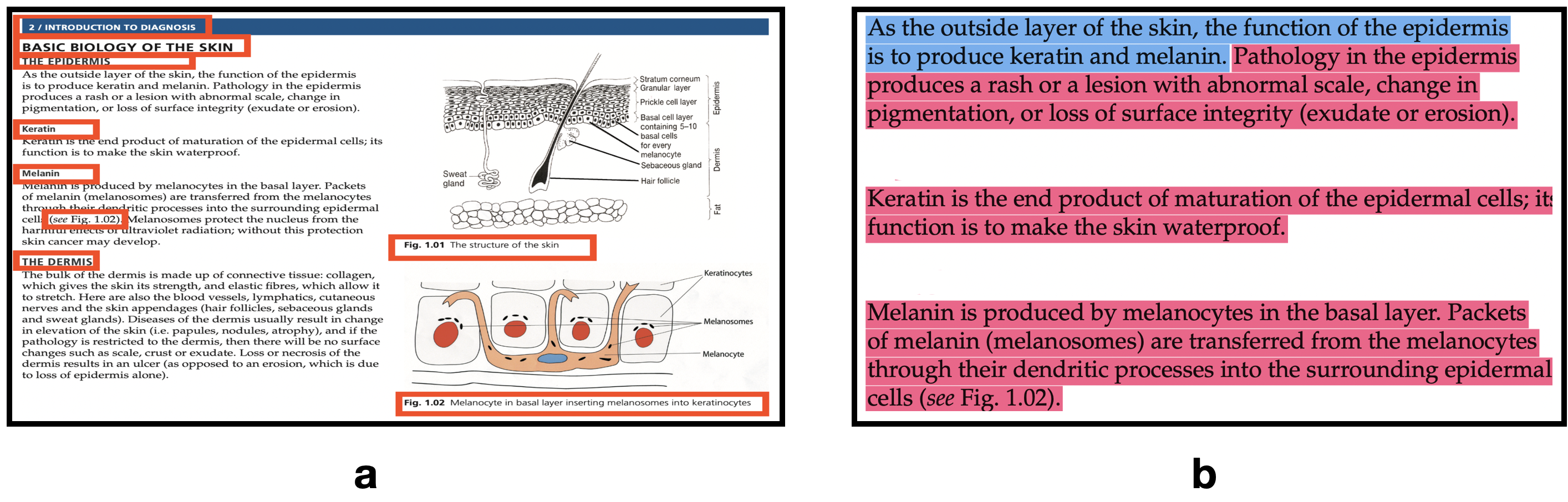}
  \caption{ \textbf{a)} Irrelevant and confounding parts of textbooks shown in red boxes are removed from the prompt-completion dataset. \textbf{b)} An example of a prompt sentence in blue with the following four sentences in pink as its completion.}
  \label{page}
\end{figure}

We knew that each dermatology book had its own structure, types of references, and formatting method. Preprocessing and extracting a proper text from the books and creating a prompt-completion dataset for further fine-tuning was divided into manual and automatic steps. The manual extraction phase was deleting irrelevant pages like glossaries, acknowledgments, and references. 
 Also, not all text in the preserved pages assisted in creating prompt-completion pairs, such as titles, footnotes, captions, tables' text, and citations. Figure~\ref{page} shows some examples. We filtered the main text by picking the lines with the dominant font and size using the PyMuPDF \footnote{https://github.com/pymupdf/PyMuPDF} python library. We assumed figures' captions or other non-informative texts like titles are less frequent in the book and have different fonts and sizes. This assumption was valid for all books we used. Table~\ref{stats} (Appendix \ref{sec:tables}) shows the names of the books and the number of prompt-completion pairs obtained for each.

% \begin{table}[h]
%   \caption{Dataset statistics}
%   \label{stats}
%   \centering
%   \begin{tabular}{||  m{4cm} | m{4cm}| m{1cm} | m {3cm} ||}
%  \hline
%  Book Name & Author(s) & Pages & Number of pairs  \\ [0.5ex] 
%  \hline\hline
%  Differential Diagnosis In Dermatology \cite{ashton2021differential} & Richard Ashton, Barbara Leppard and Hywel Cooper & 466 & 616 \\ 
%  \hline
% General Dermatology \cite{english2007general} & John SC English & 147 & 286\\
%  \hline
% Top 50 Dermatology Case Studies for Primary Care \cite{reich2017top}& Danya Reich, Corinna Eleni Psomadakis ,Bobby Buka & 341 & 851 \\
%  \hline
% Handbook of Dermoscopy \cite{malvehy2006handbook}& Josep Malvehy, Susana Puig, Ralph P Braun, Ashfaq A Marghoob, Alfred W Kopf & 106 & 58 \\
%  \hline
%  \hline
% \multicolumn{3}{|c|}{Total} & 1811 \\ [1ex] 
%  \hline
% \end{tabular}

% \end{table}

\subsection{Fine-tuning}
We fine-tuned two LLM models, GPT-2 \citep{radford2019language} and  GPT-3.5 \citep{brown2020language}, which have been pre-trained as general purpose learners on a huge amount of text data scraped from the internet. GPT-3.5 is one of the largest autoregressive language models available, trained with 4096-token-long context. However, the model is close-sourced and fine-tuning comes as part of an API endpoint. We first decided to use GPT-2, which is GPT-3.5’s predecessor with 1.5 billion parameters. To fine-tune GPT-2, we started with the extracted prompt and completion pairs from the preprocessing step. Then, we created the fine-tuning dataset by combining each prompt and completion into a single sentence separated by the padding token and tokenized the sentence using the GPT-2 tokenizer. Finally, we passed the data to the trainer with the combined prompt and completion as the label. We used the huggingface library \citep{wolf2019huggingface} to implement the GPT-2 model and fine-tuned it for two epochs. GPT-3.5 was easier to fine-tune and only needed an API key to directly call a fine-tuning endpoint. The \verb+gpt-3.5-turbo+ variant of GPT-3.5 was fine-tuned for four epochs using a similar input data from the mentioned books with the format $\{"prompt": " promptA" , "completion": "completionA" \}$.

For fine-tuning CLIP, we started by extracting the image-caption pairs from PubMed articles using the scripts provided in \cite{kim2023fostering}. We didn't use textbooks here since the repository does not have the list of textbooks used. Then, we passed the captions through the fine-tuned LLM to generated enriched captions with a max length of 512 tokens. Table~\ref{captions} (Appendix \ref{sec:tables}) shows some of the improved captions generated using fine-tuned GPT-2 and GPT-3.5 models.
 We then fine-tuned the pre-trained CLIP model \verb+openai/clip-vit-base-path32+ with a batch size of 64 using the Adam optimizer \citep{kingma2014adam} and a learning rate of 1e-5 with a cosine annealing scheduler with warm restarts.

\subsection{Zero-shot classification and Evaluation}
Once the CLIP model was fine-tuned, we used the 3230 images and corresponding concepts from the SKINCON dataset to perform zero-shot concept classification. For each concept key in the 48 SKINCON concepts, we created embeddings for the text \textit{"This is \{concept\_key\}"} and all of the images in CLIP's joint embedding space. Then, using the cosine similarity scores, we generated a Receiver operating characteristic (ROC) curve independently for each of the 48 concepts. The evaluation metric used was the area under the ROC curve (AUC).

\section{Results}
We evaluated using five different CLIP models: \textbf{1)} The vanilla CLIP model without fine-tuning (Vanilla) and the CLIP model fine tuned using \textbf{2)} Original PubMed image-caption pairs (Original). \textbf{3)} Aligned captions from fine-tuned GPT-2 \textbf{4)} Aligned captions from Vanilla GPT-3.5. \textbf{5)}  Aligned captions from fine-tuned GPT-3.5.
% \begin{enumerate}
% \item The vanilla model which is the out-of-the-box CLIP model without fine-tuning (Vanilla) 
% \item The CLIP model fine-tuned using the original PubMed image-caption pairs (Original) 
% \item The CLIP model fine-tuned using the aligned captions from fine-tuned GPT-2.
% \item The CLIP model fine-tuned using the aligned captions from Vanilla GPT-3.5.
% \item The CLIP model fine-tuned using the aligned captions from fine-tuned GPT-3.5.
% \end{enumerate}

We decided to include vanilla GPT-3.5 in our results since from qualitative analysis it seemed that GPT-3.5 by itself had a high enough expressive power to understand even techincal medical context from the captions and generate customizations. Table~\ref{mean_auc} shows the mean AUC across all concepts for the different CLIP models as defined above and Table~\ref{auc} (Appendix \ref{sec:tables}) shows the AUC scores for each of the concepts. 

\begin{table}[ht]
  \caption{Mean AUC across all concepts}
  \label{mean_auc}
  \centering
  \begin{tabular}{||  m{6cm} |  m{1cm}  ||}
 \hline
 CLIP Model & Mean AUC  \\ [0.5ex] 
 \hline\hline
 Vanilla & 0.572 \\ 
 \hline
Original & 0.636 \\
 \hline
Fine-Tuned GPT-2 & 0.642 \\
 \hline
Vanilla GPT-3.5 & 0.639 \\
 \hline
 Fine-Tuned GPT-3.5 & \textbf{0.648} \\
 \hline
 \hline
\end{tabular}

\end{table}

From Table~\ref{captions} (Appendix \ref{sec:tables}), it can be seen that the fine-tuned GPT-2 model is able to extend the input caption while keeping the sentence grammatical correct. However, it sometimes strays away from the context of the input caption and can start constructing sentences by stringing together medical jargon. This might be a result of setting a high max token length which causes the model to lose context in longer ranges. GPT-3.5 is able to maintain context for a longer token length and performs better data alignment.

Fine-tuning the CLIP model improves performance for most of the concepts (41 out of 48), see Table~\ref{auc} (Appendix \ref{sec:tables}). The fine-tuned GPT-3.5 model performs the best among all the models tested, with an AUC of 0.648 and it performs better than the original model in a majority of the concepts (26 out of 48). This indicates that fine-tuning the LLM using dermatology text helps in improving the data alignment in the extended captions.

The second best performing model is the GPT-2 fine-tuned model, with an AUC of 0.642 and performing better than the original model in 25 out of the 48 concepts. This result was unexpected since the GPT-3.5 model is much more powerful in terms of the model capacity as compared to GPT-2 and we expected the Vanilla GPT-3.5 model to outperform the fine-tuned GPT-2 model, which was not the case. This indicates that fine-tuning LLM models does actually improve the predictive performance even if the model does not have as many trainable parameters.

The Vanilla GPT-3.5 model is also able to outperform the Original model with an AUC of 0.639. This shows that LLMs can be effectively used to produce customized and well-aligned captions which improve the language supervision provided to the CLIP training procedure resulting in improved performance. 

\section{Conclusion}
Our study reveals that extending captions through the use of a fine-tuned Large Language Model (LLM) on dermatology textbooks effectively connects clinical lexicon with CLIP's pre-training data, resulting in enhanced downstream zero-shot concept classification performance in dermatology images. To summarize, our findings underscore the promise of LLMs in enhancing language supervision for dermatology AI. The improved CLIP model can be further used to annotate images with concepts that can be crucial to developing concept-based disease classifiers like concept bottleneck models \citep{koh2020concept} that are interpretable and transparent. However, further investigation is essential to optimize integration of LLMs with domain-specific models, ensuring more resilient applications in medical image analysis.

\newpage
\section{Acknowledgment}
This work was done as part of the final project for the course CSE 527 (Computational Biology) at the University of Washington. We would like to thank the professor Dr. Su-In Lee along with the teaching assistants Wei Qiu and Mingyu Lu for their valuable feedback.
\section{Limitations}
Although the early findings are promising, there are many ways to extend this project. We only used 4 dermatology textbooks for extracting the prompt-completion pairs to fine-tune the LLMs, but there are a lot more books available which can also be preprocessed. PubMed articles can be used to generate the prompt-completion pairs as well. Also, in the extraction pipeline, we made pairs by getting the following four sentences of a particular sentence without considering the context and paragraph switch. This could introduce confounders in the fine-tuning process. For instance, the first completion sentence could be related to melanoma; in contrast, the other three could be from the next section and discuss another disease. In addition, python pdf parsers occasionally fail and break some words into meaningless chunks that can doubtlessly mislead the LLM during fine-tuning. A solution for the extraction issues is adding more manual and automatic steps to remove and filter meaningless words and checking the context integration. LLMs have also been known to hallucinate \citep{lee2018hallucinations, bang2023multitask} and proper steps need to be taken to ensure non-existent facts are not fabricated which is pertinent in a high-stakes domain like dermatology. 

Furthermore, we used the \verb+gpt-3.5-turbo+ variant of GPT-3.5, but there are more powerful variants available like GPT-4 which we did not use due to budget constraints. Another approach to enhance the performance of the fine-tuned Large Language Model (LLM) and refine the generated captions is by incorporating Instruction Tuning data \citep{zhang2023instruction,liu2023visual, dai2023instructblip, ouyang2022training}, instruction-output pairs, extracted from dermatology books during the fine-tuning process. This task needs a careful plan to create a dataset that is useful and gives valuable insights.

Another change that could be made is the CLIP model used. We used the \verb+openai/clip-vit-base-path32+ model for CLIP training but there is a more powerful baseline CLIP model \verb+openai/clip-vit-large-patch14+ available, which we did not use because of memory constraints and longer training times. We can also employ a non-random batch sampling strategy, which includes samples with different concepts in one mini-batch for efficient learning of concepts. Another way to improve language supervision by employing ways to increase the number of tokens from 77, which is CLIP's limitations. We anticipate that all of these changes will improve the zero-shot classification performance of the fine-tuned CLIP model.

\clearpage
\bibliography{bibliography}
\bibliographystyle{iclr2024_conference}
\clearpage

% \newpage
\appendix
\section{Appendix}
\subsection{Related Work}
The CLIP network \citep{radford2021learning} learns visual concepts by being trained with image and text pairs in a self-supervised manner, using text paired with images found across the Internet. CLIP uses a contrastive learning procedure to generate a multi-model embedding space by jointly training an image encoder and a text encoders such that the embeddings of a given image-text pair are close together in the joint representation space. Given a batch of $N$ (image, text) pairs, CLIP is trained to predict which of the $N \times N$ possible (image, text) pairings across a batch actually occurred. This is done by maximizing the cosine similarity of the image and text embeddings of the $N$ real pairs in the batch while minimizing the cosine similarity of the embeddings of the $N^2 - N$ incorrect pairings. This optimization is done using a symmetric cross entropy loss over these similarity scores. CLIP is powerful enough to be used in zero-shot manner on standard images (such as those from ImageNet \citep{deng2009imagenet} classes). However, dermatology images are sufficiently different from everyday images that it would be useful to fine-tune CLIP with them. 

There has been prior work done for performing self-supervised constrastice learning tasks in the medical domain. \citep{tiu2022expert} used contrastive learning for training a self-supervised model on chest x-ray images lacking explicit annotations to perform pathology-classification tasks. However, the MIMIC-CXR dataset \citep{johnson2019mimic} which was used to train the model consists of expert radiology reports accompanying each image which has rich textual descriptions about the x-ray and enables the text transformer to better learn visual medical concepts and generalize to different pathologies. In case of dermatology images, no such dataset exists containing images with corresponding expert reports. 

% \citep{chaitanya2020contrastive} introduces novel contrasting strategies and a local version of the contrastive loss to perform per-pixel segmentation on three Magnetic Resonance Imaging (MRI) datasets. However, they still use limited annotations to learn global and local features from the MRI scans.

In the language supervision domain, several prior works have used text-based knowledge of image categories to improve classification accuracy. \cite{elhoseiny2017link} extracts visual information from unstructured text descriptions collected from
the internet to recognize object parts and perform zero-shot classification. \cite{paz2020zest} extract visual information from Wikipedia descriptions to enable bird
classification. These works show that text augmentation is useful for self-supervised models, but they rely on external natural language datasets for generating descriptions. \cite{pratt2022does} uses a large language model (GPT-3.5) to generate customized prompts for a range of zero-shot
image classification benchmarks, however it does not perform any fine-tuning and the datasets don't include any images from the medical domain.

Our proposed approach combines the prior works by leveraging the expressive power provided by LLMs with the availability of dermatology images and contrastive learning approaches to improve zero-shot concept generation using customized prompts. 

% \paragraph{PubMed Articles} Training the CLIP model requires a dataset which consists of image and text pairs used for the contrastive learning procedure. We used PubMed articles which provide a rich source of images and corresponding captions. The work of extracting of image caption pairs from the PubMed articles has been previously done in \cite{kim2023fostering}. Free access PubMed articles were collected and filtered using the search terms \textit{dermatology}, \textit{dermoscopy}, \textit{melanoma}, and \textit{skin disease}. After obtaining all the images, the images which were not related to dermatology (like tables and figures) were filtered out using a similarity score function to a known dermatology image, resulting in a total of 44314 image-caption pairs. The captions were then extended by passing them through the fine-tuned LLM and finally used to train the CLIP model. 

\newpage
\subsection{Tables}
\label{sec:tables}

\begin{table}[h]
  \caption{Tokenized caption statistics}
  \label{cap_len_stats}
  \centering
  \begin{tabular}{||  m{6cm} |  m{1cm}  ||}
 \hline
 Statistic & Value  \\ [0.5ex] 
 \hline\hline
 Mean & 35.394 \\ 
 \hline
Standard Deviation & 22.801\\
 \hline
Minimum & 3 \\
 \hline
Lower Quartile & 17 \\
 \hline
 Median & 28 \\
 \hline
 Upper Quartile & 51 \\
 \hline
 Maximum & 77 \\
 \hline
 \hline
\end{tabular}

\end{table}

\begin{table}[h]
  \caption{Dataset statistics}
  \label{stats}
  \centering
  \begin{tabular}{||  m{6cm} |  m{1cm} | m {3cm} ||}
 \hline
 Book Name & Pages & Number of pairs  \\ [0.5ex] 
 \hline\hline
 Differential Diagnosis In Dermatology \citep{ashton2021differential} & 466 & 616 \\ 
 \hline
General Dermatology \citep{english2007general} & 147 & 286\\
 \hline
Top 50 Dermatology Case Studies for Primary Care \citep{reich2017top}&  341 & 851 \\
 \hline
Handbook of Dermoscopy \citep{malvehy2006handbook}&  106 & 58 \\
 \hline
 \hline
\multicolumn{2}{|c|}{Total} & 1811 \\ [1ex] 
 \hline
\end{tabular}

\end{table}

\begin{table}[h]
  \caption{Examples of improved captions using fine-tuned GPT-2 and GPT-3.5}
  \label{captions}
  \centering
  \begin{tabular}{||  m{3cm} |  m{5cm} | m{5cm}||}
 \hline
 Caption & GPT-2 Fine-Tuned  &  GPT-3.5 Fine-Tuned\\ 
 \hline\hline
 \texttt{Long-standing disseminated granuloma annulare on the left leg.} & \texttt{Infection with streptococci can cause pustular psoriasis on the lower legs, a condition which is commonly referred to as “sore throat on the legs.” Granuloma annulare is a common, self-limiting, skin-tearing condition. It occurs in about 10 \% of immunocompromised populations and occurs predominantly at sites where skin is heavily covered, such as the arms and legs.'}  & \texttt{This long-standing disseminated granuloma annulare (DGA) on the left leg is likely caused by an unknown infection. The DGA typically presents with a red, inflamed lump that is hard to see. It can cause significant pain and difficulty walking. Treatment typically involves antibiotics and/or surgery.}\\ 
 \hline
 \texttt{Anakinra-induced sarcoidosis. Buttock erythematous plaque eruption while on anakinra for hidradenitis suppurativa.} & \texttt{Multiple papillary lesions scattered symmetrically over the lower abdominal area with or without a border, sometimes with or without scaling. Itching on palpation.The clinical appearance and pathogenesis of sarcoid vary greatly depending on the size, location and age of the patient. In the elderly, it is typically chronic, and lasts for many years.} & \texttt{Anakinra is a medication used to treat certain inflammatory conditions, including hidradenitis suppurativa. In rare cases, anakinra has been linked to an increased risk of developing sarcoidosis, a condition that causes abnormal collections of inflammatory cells in the body. Signs and symptoms of sarcoidosis can vary but can include skin rashes and lesions, such as an erythematous plaque eruption on the buttocks.}\\
 \hline
 \texttt{Interdigital furunculosis with a painful, erythematous and alopecic nodule.} & 
 \texttt{These frequently rupture to drain a haemopurulent discharge. Individual lesions may be often caused by penetrating foreign bodies, but recurrent lesions are more commonly associated with atopic dermatitis and/or conformational problems.The discharge is red, erythematous and itchy. It gets better spontaneously after few days to a week.} & \texttt{Atopic eczema tends to be poorly defined, but with prominent warty plaques. In peridial eczema the surface is dry and erythematous, and is complicated by eosinophilia and by chronic, unresponsive or nodular form.The differential diagnosis of peridial eczema includes solar eczema, rubella and psoriasis may turn  from pink to deep purple with the development of scabs.}\\
\hline
\end{tabular}
\end{table}

\begin{table}[h]
  \caption{Concept Classification AUC. \# True shows the number of samples in the SKINCON dataset labeled with that concept}
  \label{auc}
  \centering
  \begin{tabular}{||  m{5cm} | m{1cm} | m{1cm} | m{1cm} | m {1cm} | m {1cm} | m{1cm} ||}
 \hline
 Concept & \# True & Vanilla & Original & Fine-Tuned GPT-2 & Vanilla GPT-3.5 &  Fine-Tuned GPT-3.5\\ \hline
        Vesicle & 46 & 0.552 & 0.667 & \textbf{0.715} & 0.674 & 0.584 \\ \hline
        Papule & 1170 & 0.448 & 0.66 & 0.646 & 0.668 & \textbf{0.676} \\ \hline
        Macule & 13 & 0.407 & 0.395 & 0.54 & 0.469 & \textbf{0.556} \\ \hline
        Plaque & 1967 & 0.566 & 0.646 & 0.582 & \textbf{0.656} & 0.611 \\ \hline
        Abscess & 5 & \textbf{0.929} & 0.847 & 0.787 & 0.875 & 0.884 \\ \hline
        Bulla & 64 & 0.508 & 0.611 & 0.584 & 0.647 & \textbf{0.654} \\ \hline
        Patch & 149 & 0.547 & 0.461 & 0.546 & \textbf{0.599} & 0.523 \\ \hline
        Nodule & 189 & 0.719 & \textbf{0.773} & 0.744 & 0.758 & 0.758 \\ \hline
        Ulcer & 154 & 0.82 & 0.883 & \textbf{0.886} & 0.879 & 0.883 \\ \hline
        Crust & 497 & 0.559 & 0.666 & 0.635 & 0.671 & \textbf{0.727} \\ \hline
        Erosion & 200 & 0.538 & 0.593 & \textbf{0.626} & 0.603 & 0.602 \\ \hline
        Excoriation & 46 & 0.536 & \textbf{0.693} & 0.6 & 0.559 & 0.578 \\ \hline
        Atrophy & 69 & 0.482 & 0.606 & 0.613 & 0.563 & \textbf{0.616} \\ \hline
        Exudate & 144 & \textbf{0.677} & 0.656 & 0.617 & 0.629 & 0.626 \\ \hline
        Purpura/Petechiae & 10 & 0.577 & 0.592 & 0.662 & \textbf{0.667} & 0.646 \\ \hline
        Fissure & 32 & 0.708 & 0.548 & 0.428 & 0.506 & \textbf{0.686} \\ \hline
        Induration & 33 & \textbf{0.594} & 0.559 & 0.528 & 0.573 & 0.553 \\ \hline
        Xerosis & 35 & 0.41 & 0.735 & 0.737 & \textbf{0.744} & 0.547 \\ \hline
        Telangiectasia & 100 & 0.366 & 0.484 & \textbf{0.574} & 0.47 & 0.564 \\ \hline
        Scale & 686 & 0.485 & 0.474 & 0.434 & 0.417 & \textbf{0.521} \\ \hline
        Scar & 123 & 0.604 & \textbf{0.659} & 0.592 & 0.568 & 0.639 \\ \hline
        Friable & 153 & \textbf{0.629} & 0.576 & 0.628 & 0.555 & 0.377 \\ \hline
        Sclerosis & 27 & \textbf{0.661} & 0.557 & 0.582 & 0.595 & 0.506 \\ \hline
        Pedunculated & 26 & 0.665 & \textbf{0.855} & 0.755 & 0.817 & 0.773 \\ \hline
        Exophytic/Fungating & 42 & \textbf{0.713} & 0.629 & 0.657 & 0.607 & 0.7 \\ \hline
        Warty/Papillomatous & 46 & \textbf{0.71} & 0.591 & 0.592 & 0.636 & 0.691 \\ \hline
        Dome-shaped & 146 & 0.624 & 0.604 & \textbf{0.71} & 0.658 & 0.667 \\ \hline
        Flat topped & 18 & 0.574 & 0.595 & 0.609 & 0.563 & \textbf{0.635} \\ \hline
        Brown(Hyperpigmentation) & 760 & 0.648 & 0.768 & \textbf{0.776} & 0.763 & 0.738 \\ \hline
        Translucent & 16 & 0.496 & 0.523 & 0.69 & \textbf{0.731} & 0.547 \\ \hline
        White(Hypopigmentation) & 257 & 0.596 & 0.686 & 0.718 & 0.715 & \textbf{0.737} \\ \hline
        Purple & 85 & 0.725 & \textbf{0.843} & 0.813 & 0.777 & 0.762 \\ \hline
        Yellow & 245 & 0.614 & \textbf{0.744} & 0.733 & 0.706 & 0.721 \\ \hline
        Black & 90 & 0.685 & 0.873 & \textbf{0.901} & 0.882 & 0.896 \\ \hline
        Erythema & 2139 & 0.609 & \textbf{0.719} & 0.711 & 0.666 & 0.68 \\ \hline
        Comedo & 24 & 0.469 & 0.502 & 0.527 & 0.561 & \textbf{0.632} \\ \hline
        Lichenification & 25 & 0.505 & \textbf{0.565} & 0.55 & 0.545 & 0.502 \\ \hline
        Blue & 5 & 0.662 & 0.749 & 0.767 & 0.754 & \textbf{0.784} \\ \hline
        Umbilicated & 49 & 0.57 & 0.683 & 0.567 & 0.663 & \textbf{0.751} \\ \hline
        Poikiloderma & 5 & 0.324 & \textbf{0.621} & 0.453 & 0.4 & 0.524 \\ \hline
        Salmon & 10 & 0.463 & \textbf{0.671} & 0.641 & 0.667 & 0.588 \\ \hline
        Wheal & 21 & 0.507 & \textbf{0.796} & 0.775 & 0.666 & 0.693 \\ \hline
        Acuminate & 8 & 0.444 & 0.279 & 0.588 & \textbf{0.654} & 0.606 \\ \hline
        Burrow & 5 & \textbf{0.807} & 0.636 & 0.585 & 0.68 & 0.786 \\ \hline
        Gray & 5 & 0.302 & 0.45 & 0.439 & 0.283 & \textbf{0.303} \\ \hline
        Pigmented & 5 & 0.459 & 0.483 & 0.513 & 0.581 & \textbf{0.661} \\ \hline
        Cyst & 6 & 0.521 & 0.745 & \textbf{0.883} & 0.79 & 0.827 \\ \hline
 \hline
 \hline
\multicolumn{2}{|c|}{Mean AUC} & 0.572 & 0.636 & 0.642 & 0.639 & \textbf{0.648} \\ [1ex] 
 \hline
\end{tabular}

\end{table}

\end{document}